\documentclass[journal]{IEEEtran}
\usepackage{graphicx}
\usepackage[bf]{caption}
\usepackage{cite} 
\usepackage{svg}
\usepackage{url}
\usepackage{amssymb}    
\usepackage{amsmath}
\usepackage{mathtools}          
\usepackage[hidelinks]{hyperref}
\graphicspath{{./img/}}
\newcommand{\RR}{\mathbb{R}}
\newcommand{\B}[1]{\mathbf{#1}} 

\DeclareCaptionLabelFormat{custom}
{%
    #1 \textbf{#2.}
}

\begin{document}

\title{Closed-form control with spike coding networks}

\author{Filip~S.~Slijkhuis,
        Sander~W.~Keemink$^*$,
        and~Pablo~Lanillos$^*$\\
        $^*$These authors contributed equally
\thanks{The authors are with the Donders Institute for Brain, Cognition and Behavior. Department of AI, Radboud University, Nijmegen, the Netherlands. e-mail: filipslijkhuis[at]outlook.com, \{sander.keemink, p.lanillos\}[at]donders.ru.nl}
\thanks{P.L. is partially supported by the Spikeference project, Human Brain Project Specific Grant Agreement 3 (ID: 945539).}
}



\markboth{}%
{Slijkhuis\MakeLowercase{\textit{et al.}}: Closed-form control with spike coding networks}

\maketitle

\begin{abstract} 
Efficient and robust control using spiking neural networks (SNNs) is still an open problem. Whilst behaviour of biological agents is produced through sparse and irregular spiking patterns, which provide both robust and efficient control, the activity patterns in most artificial spiking neural networks used for control are dense and regular --- resulting in potentially less efficient codes. Additionally, for most existing control solutions network training or optimization is necessary, even for fully identified systems, complicating their implementation in on-chip low-power solutions. The neuroscience theory of Spike Coding Networks (SCNs) offers a fully analytical solution for implementing dynamical systems in recurrent spiking neural networks --- while maintaining irregular, sparse, and robust spiking activity --- but it's not clear how to directly apply it to control problems. Here, we extend SCN theory by incorporating closed-form optimal estimation and control. The resulting networks work as a spiking equivalent of a linear–quadratic–Gaussian controller. We demonstrate robust spiking control of simulated spring-mass-damper and cart-pole systems, in the face of several perturbations, including input- and system-noise, system disturbances, and neural silencing. As our approach does not need learning or optimization, it offers opportunities for deploying fast and efficient task-specific on-chip spiking controllers with biologically realistic activity.

\end{abstract}

\begin{IEEEkeywords}
    Spiking Neural Networks, State Estimation, Optimal Control, Spike Coding Networks, Dynamical Systems.
\end{IEEEkeywords}
 
\section{Introduction}

Brain and behaviour are inseparable. The activity of complex networks of neurons is strongly linked to the capacity of biological agents to move and interact in the world. These networks control the body through sparse spiking activity~\cite{gerstner_neuronal_2014,abbott_building_2016}, providing high energy-efficiency and robustness against perturbations (e.g. noise or neural silencing)~\cite{deneve_brain_2017,leary_annual_2003,morrison_life_1997}. Whilst SNNs are experiencing a qualitative improvement in recent years driven by neuromorphic hardware and learning algorithms developments~~\cite{davies_advancing_2021,neftci_surrogate_2019}, there are still open challenges in applying them to control problems. 

First, the majority of the SNN-based solutions rely on training or otherwise optimizing model parameters \cite{bing_survey_2018,traub_many-joint_2021,davies_advancing_2021}, even for fully identified systems. Analytical solutions are desirable for control, as they are interpretable, have the advantage of being well-applicable to identified systems, are quick and efficient to implement and deploy, and they are amendable to theoretical explorations of stability and function. In control theory, such fully analytical solutions are often possible, but are difficult to directly implement in SNNs due to their highly nonlinear and discontinuous nature.

Second, biological spiking codes are generally highly irregular and sparse \cite{tolhurst_statistical_1983, shadlen_variable_1998} --- and require little energy. In contrast, SNN-based control solutions often produce and require highly dense and regular activity (e.g., \cite{eliasmith_large-scale_2012,salaj_spike_2021}), and are hence inefficient~\cite{hubotter_training_2021}. How we can reconcile precise and efficient control with a more biological irregular spiking code is an open problem. 

While spiking irregularity is usually considered a consequence of noise, according to the neuroscience theory of spike coding networks (SCNs) it could also be a signature of a highly precise spiking code. 
SCN theory follows a similar principle as the theory of predictive coding. Neurons only fire when the network's prediction error exceeds a threshold value, efficiently constraining this error. The resulting neural spiking activity is `coordinated' across the neuron population, producing sparse and irregular patterns  \cite{boerlin_predictive_2013,calaim_geometry_2022}, and is highly robust against several biological perturbations~\cite{barrett_david_optimal_2016,calaim_geometry_2022}. SCNs have the great advantage that they are defined analytically --- through a closed-form solution for the recurrent connectivity. They thus have the potential to solve both above challenges. However, while the SCN framework permits us to analytically implement~\cite{boerlin_predictive_2013,nardin_nonlinear_2021} or learn~\cite{alemi_learning_2018,brendel_learning_2020,thalmeier_learning_2016} any dynamical system, it is an open problem how they can be used to both estimate and control the state of an external system. 

\begin{figure*}[!t]
    \centering
    \includegraphics[width=0.9\linewidth]{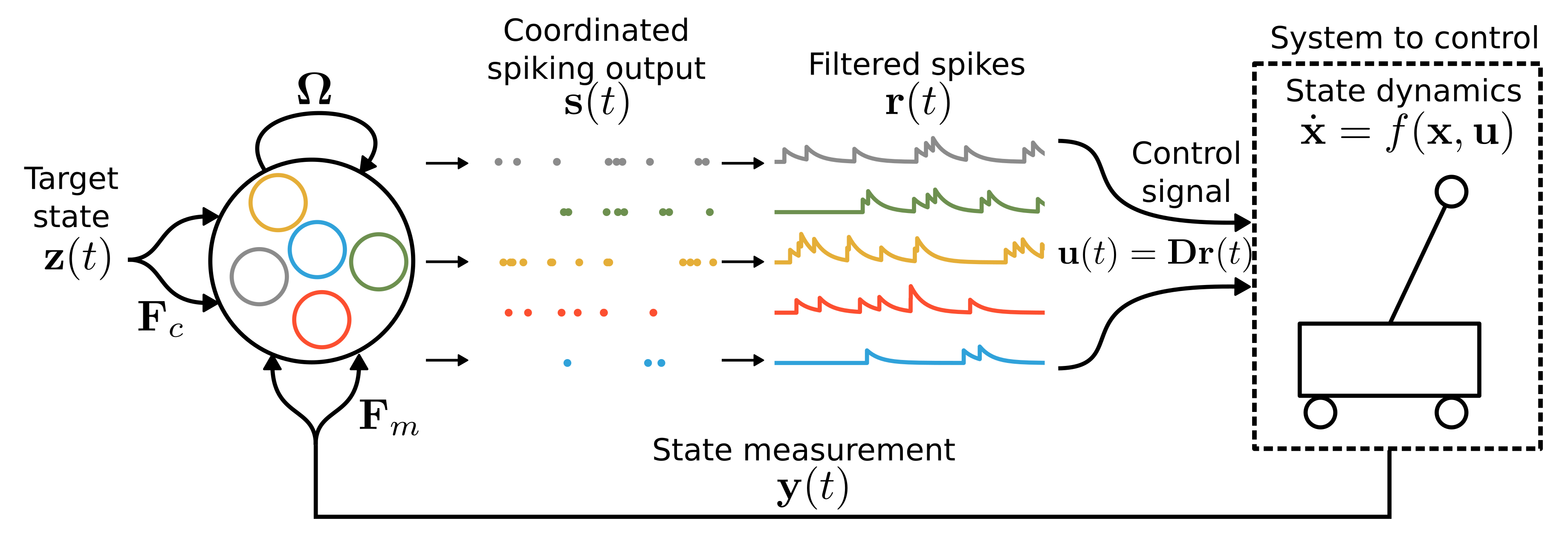}
    \caption{Controlling dynamical systems with spiking neural networks. A recurrent spiking neural network (left) will emit spikes based on its inputs and connectivity, which are translated into a control signal to control some target system (right). $\B F_c$ and $\B F_m$ represent input weights, $\B \Omega$ represent recurrent weights, and $\B D$ represent read-out weights. By first defining the optimal control solution for the system, and then directly translating the resulting control parameters into these connection weights, we can generate a robust and population-wide coordinated spiking network that accurately controls identified systems.}
    \label{fig:intro}
\vspace{-1em}
\end{figure*}

In this paper we combine optimal control and SCN theory to address both of the above challenges --- providing a closed-form solution for optimal control using spiking neural networks, given a well identified system, while producing realistic and robust spiking patterns~(Fig.~\ref{fig:intro}).  Our proposed method for estimation and control both expands on a promising theory for understanding biological spiking activity, as well as provides a major step towards developing low-power, high-efficiency, robust, and task-specific neuromorphic controllers.


\subsection{Contribution}
We mathematically formalize the link between optimal control of dynamical systems and the SCN framework. Particularly, we analytically derive the spiking equivalent of the linear–quadratic–Gaussian (LQG) control problem -- the combination of a linear quadratic estimator (Kalman filter) and a linear–quadratic regulator (LQR) controller (Fig.~\ref{fig:1}B and C). We show that our proposed networks ($i$) accurately estimate and control well-known dynamical systems, achieving similar performance as their non-spiking counterpart, and, importantly, ($ii$) they preserve the irregular and sparse spiking patterns and robustness to neural silencing.

Developing efficient spiking neural controllers with fully derivable and interpretable connectivity is especially relevant for industrial applications~\cite{davies_advancing_2021}, robotics~\cite{traub_many-joint_2021} and machine intelligence in general~\cite{eliasmith_large-scale_2012}. 
 
\subsection{Related SCN literature}
Here, we non-exhaustively review the main lines of research that followed a similar approach as us. For recent reviews on SNNs related to robotics and control see~\cite{bing_survey_2018,sandamirskaya_neuromorphic_2022}. Our work is based on~\cite{boerlin_predictive_2013}, which originally showed how any \textbf{linear} dynamical system can be analytically implemented in an SCN. While there have been several follow-up papers on extending the framework for general computations \cite{alemi_learning_2018,mancoo_understanding_2020,nardin_nonlinear_2021}, control has been less well-studied. In \cite{thalmeier_learning_2016}, spike coding networks were extended by adding learning rules for connectivity weights. This allowed them to perform forward prediction of the dynamical system state. As an additional result, they showed how to control a pendulum by using the network to simulate multiple future trajectories of the pendulum with different control policies. However, the control algorithm is not encoded within the network and does not provide robust estimation. In \cite{huang_optimizing_2017, huang_dynamical_2018}, the authors proposed an analytical SCN-inspired framework for control. They mainly focused on producing the correct control signal, and derived network connectivities accordingly. The framework requires fully observable systems, and they were not able to provide a general, clear and simple mathematical model that is in line with both control and neuroscience communities. Furthermore, they did not investigate the beneficial properties of the spiking patterns generated by the coordination of the neurons. Here, we will provide a simpler analytical derivation more in line with existing SCN theory, derive networks for both estimation and control of partially observable systems, and investigate the robustness properties of the control and networks in more detail. Overall, our approach takes inspiration from the broader SCN literature to provide a unified mathematical framework to analytically compute optimal control in SNNs.

\section{Methods}

\subsection{The spiking control problem}
\label{sec:methods:scn}

We have a system with state $\B x(t)$ we would like to control with a spiking signal ($\B s(t)$) emitted from a recurrent SNN~(Fig.~\ref{fig:intro}). The SNN is provided with some incomplete measurements of the system state $\B y(t)$ and a target state $\B z(t)$, and has to generate a control signal $\B u(t)$. 


The spiking patterns are generated according to some underlying voltage dynamics. There are many such models of varying complexities, but for most practical applications networks of leaky integrate-and-fire (LIF) neurons suffice. A network of $N$ such neurons is then defined by 
\begin{equation}\label{general_LIF}
    \B{\dot{v}}(t) = -\lambda\B{v}(t) + \B{F}\B c(t) + \B{\Omega_s \B r}(t) + \B{\Omega_f s}(t)  + \B \eta_v \text{,}
\end{equation}
where $\B v(t) \in \RR^N$ is the vector of neural voltages, $\B F \in \RR^{K\times N}$ are some input weights, $\B c(t) \in \RR^K$ is some $K$-dimensional input, $\B s(t) \in \RR^N$ are the emitted spikes, $\B r(t) \in \RR^N$ are the filtered spike-trains\footnote{such that $\dot {\B r} = -\lambda \B r + \B s$, see Fig.~\ref{fig:intro}.}, and $\B \eta_v \in \RR^N$ corresponds to independent voltage noise. Finally, there are fast synapses which affect the post-synaptic voltage instantaneously following a spike (through fast recurrent connections $\B \Omega_f \in \RR^{N\times N}$), and slow synapses which cause an initially slow change in voltage  (through the slow recurrent connections $\B \Omega_s \in \RR^{N\times N}$).

Whenever a given neuron's voltage $v_i$ reaches a threshold $T_i$, that neuron will emit a spike at that time (resulting in spike-trains $s_i(t)=\sum_j \delta(t-t_j)$). The voltage will then be reset to some resting value (here implemented through the diagonal elements of $\B \Omega$). 

How should the recurrent and input weights of the network be set, such that the output signal is as optimal as possible? One solution is to train the network using a cost function that enforces, for instance, sparsity~\cite{hubotter_training_2021}. Here we take another route by first defining the solution according to classical control theory, and then translating this solution into the spiking network parameters by using the neuroscience theory of Spike Coding Networks (SCN)~\cite{boerlin_predictive_2013, deneve_efficient_2016}. 


\begin{figure*}[hbtp!]
     \centering
        \includegraphics[width=0.8\textwidth]{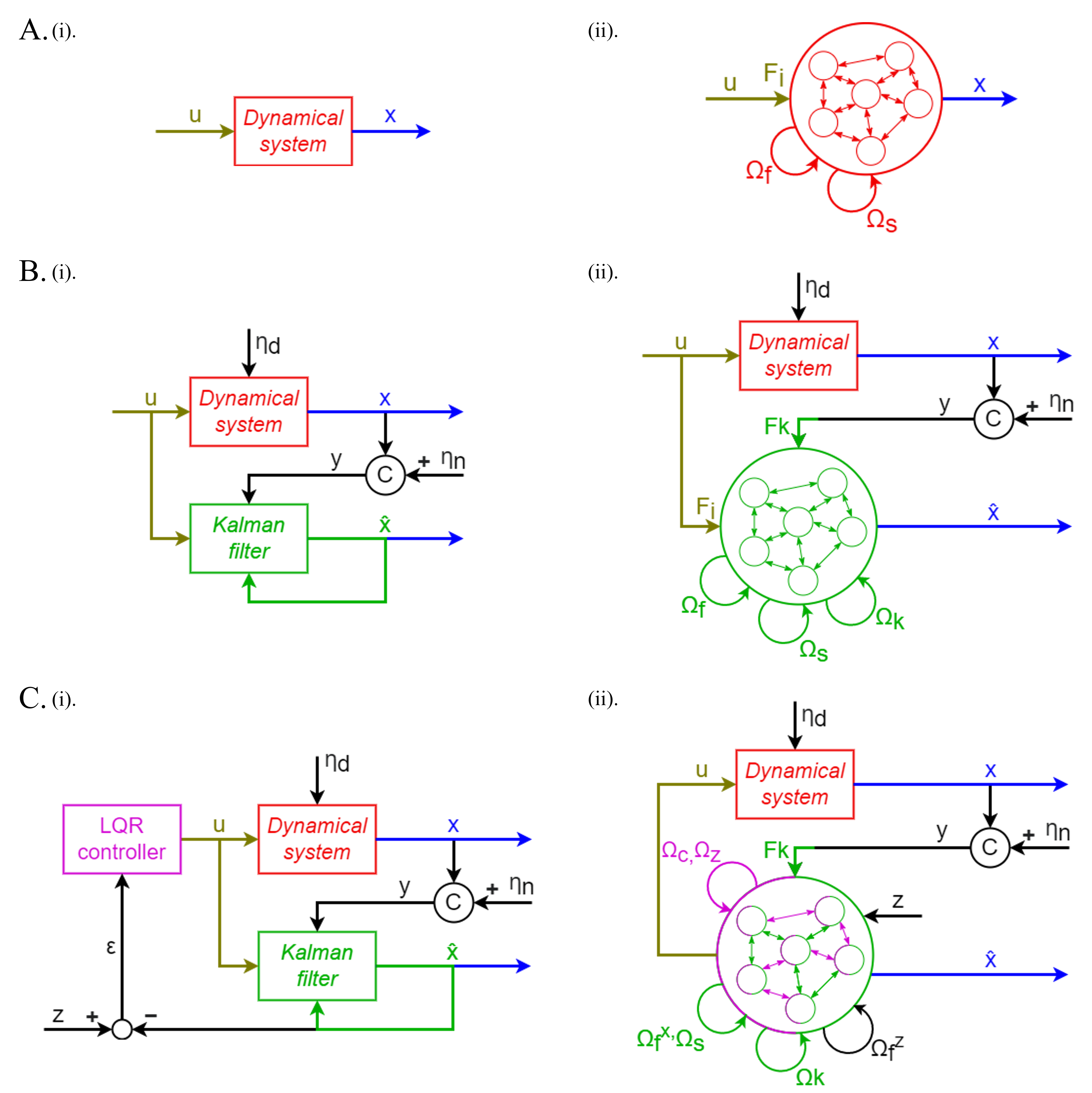}
        \caption{Schematics for identification, estimation and control of dynamical systems for classical control (left column) and its spiking neural network equivalent (right column). The internal recurrent connections in the neural representations are illustrative---see Sec~\ref{sec:methods:model} for more details. \textbf{(A)} Identification of dynamical systems~\cite{barrett_david_optimal_2016,nardin_nonlinear_2021}. This work: \textbf{(B)} Optimal estimation of dynamical systems through Kalman filtering. \textbf{(C)} Optimal estimation and control of dynamical systems through Kalman filtering combined with an LQR controller.}
        \label{fig:1}
\vspace{-1em}
\end{figure*}


\subsection{Spike-coding network theory}
For the sake of completeness, in this section we will give a brief overview of the original SCN framework (as originally proposed in \cite{boerlin_predictive_2013}) for implementing linear dynamical systems in an SNN --- for an extensive mathematical derivation of the SCN we refer the reader to~\cite{barrett_david_optimal_2016,nardin_nonlinear_2021}. 

\subsubsection{Tracking a fully observable system}
If a $K$-dimensional state $\B{x}(t) \in \mathbb{R}^{K}$ is fully observable without noise, SCN theory defines how a recurrent SNN can optimally track this signal~\cite{boerlin_predictive_2013}. The proof starts with two assumptions: 1) the input signal estimate $\B{\hat{x}}$ can be linearly decoded from the spiking activity as $\B{\hat{x}} = \B{Dr}$, where $\B D \in \mathbb{R}^{N\times K}$ are known decoding weights (in this paper randomly drawn from the normal distribution and normalized); and 2) spikes should only be emitted when this improves a coding error defined by the L2-norm, yielding a greedy spiking rule, i.e. neuron $i$ should spike \textit{iff} $||\B x - \B D \B r||^2_2 > ||\B x - \B D \B r - \B D_i||^2_2$. Here $\B D_i $ is the $i$'th column of $\B D$, and reflects the change in error due to neuron $i$ spiking.

From these assumptions a recurrent network of leaky integrate-and-fire neurons is directly derived of the form
\begin{equation}\label{derivation_encoder}
    \B{\dot{v}} = -\lambda\B{v} + \B{D}^\top(\B{\dot{x}} + \lambda\B{x}) + \boldsymbol \Omega_f \B s \text{.}
\end{equation}
In this network, the input $\B{x}$ is encoded through forward weights $\B D^\top$, with the derivative term $\dot{\B x}$ ensuring that quick changes in $\B x$ are adequately tracked. Effectively, the network takes in both the current state and its dynamics as inputs. The recurrent connections are given by fast connections $\boldsymbol \Omega_f = -\B D^\top \B D$. These connections make sure that the spiking in the network is coordinated across the neurons, such that there are no superfluous spikes. A neural post-spike `self-reset' is implicitly included by $\boldsymbol \Omega_f$'s diagonal. Neurons will emit a spike when their threshold is hit, which follows from the same derivation as $T_i=\B D_i^\top \B D_i/2$. We here assume instantaneous communication between neurons, such that only one neuron spikes at a given time (since they can instantly inhibit each-other), but note that this assumption is not strictly required\cite{calaim_geometry_2022}. The resulting spiking activity will optimally encode the input-signal such that $\hat{\B x}(t) = \B D \B r(t) \approx \B x (t)$.

\subsubsection{Implementing a dynamical system}\label{sec:dynimpl}
Instead of tracking an external signal ($\hat{\B{x}} \approx \B{x}$) one can also use SCN theory to define an SNN which implements a linear dynamical system of the form $\dot{\B x} = \B A \B x$~\cite{boerlin_predictive_2013} (Fig.~\ref{fig:1}A). Briefly, we can replace $\dot{\B{x}}$ (in Eq.~\ref{derivation_encoder}) by the now \textbf{known} dynamics $\B{A} \B{x}$. The resulting voltage dynamics are given by

\begin{align}\label{derivation_with_dynamics_1}
    \B{\dot{v}} & = -\lambda\B{v} + \B{D}^\top(\B{Ax} + \lambda\B{x}) - \B{D}^\top\B{Ds} + \eta_{V} \text{.}
\end{align}
By next replacing the external state $\B{x}$ by the network's internal estimate $\hat{\B{x}} = \B{Dr}$ this can be further simplified as
\begin{align}\label{derivation_with_dynamics_2}
    \B{\dot{v}} & = -\lambda\B{v} + \B{\Omega}_s\B{r} + \B{\Omega}_f\B{s} + \eta_{V} \text{,}
\end{align}

where the derivation has resulted in an additional set of slow connections implementing the desired dynamics given by $\B{\Omega}_s = \B{D}^\top(\B{A}+\lambda\B{I})\B{D}$. In essence, the slow connectivity reads out the internal estimate of $\B x$ through $\B{D}$. It then applies the dynamics of the linear dynamical system through $\B{A} + \lambda \B I$, and encodes the result back into the network using $\B{D}^{\top}$. The network now keeps track of its own estimate on a fast time scale through the fast connections, and drifts this estimate according to the dynamics $\B A$ through the slow recurrent connections. 

\section{Closed-form optimal estimation and control with SNNs}
\label{sec:methods:model}
In the previous section, we summarized how SCNs can implement dynamical systems --- but how can we link such a network to an external system and control it? Here we provide a new mathematical framework which extends SCN theory to allow the implementation of classical control in SNNs consisting of leaky integrate-and-fire neurons. The inner workings of the networks, and the interpretation of the different connectivities, remain unchanged from the previous section. The resulting recurrent SNN is able to simultaneously act as a state-estimator (Kalman filter) and a controller (LQR) for an external dynamical system. Precisely, we provide a linear–quadratic–Gaussian spiking controller with extended benefits of efficient spiking patterns coding.

\subsection{The classical control problem}
Consider a $K$-dimensional linear dynamical system in state-space representation (Fig.~\ref{fig:1}A.i):
\begin{align}\label{eq:dynsystem_x}
    \B{\dot{x}}(t) &= \B{A}\B{x}(t) + \B{B}\B{u}(t) + \boldsymbol{\eta}_{d} \\
    \B{y}(t) &= \B{C}\B{x}(t) + \boldsymbol{\eta}_{n}\text{,}
\end{align}
where $\B{x}(t) \in \RR^{K}$ is the state vector, $\B{A} \in \RR^{K \times K}$ is the system matrix conveying the dynamics of the system, $\B{u}(t) \in \mathbb{R}^{P}$ is the input vector through which we can control the system, and $\B{B} \in \RR^{K \times P}$ is the input matrix. $\boldsymbol \eta_{d} \in \RR^K$ reflects internal disturbances, given by a zero-mean Gaussian process with co-variance $\B{\Sigma}_{d}$. $\B y(t) \in \RR^Q$ are observations about $\B{x}(t)$, where $\B{C} \in \RR^{Q \times K}$ is the observability matrix. $\boldsymbol \eta_{n}$ is sensor noise, given by a zero-mean Gaussian process with co-variance $\B{\Sigma}_{n}$. 

We will generally assume as known the measurement vector $\B{y}$, the system matrix $\B{A}$, input matrix $\B{B}$, the input vector $\B{u}$, the observability matrix $\B{C}$, and co-variances $\B{\Sigma}_{d}$ and $\B{\Sigma}_{n}$.

Our goal is to control $\B x$ to some reference state $\B{z} \in \mathbb{R}^{K}$. To do this we must be able to estimate the state $\B{x}$ from the observations $\B{y}$ (\textit{estimation problem}), and find the best control signal $\B{u}$ to do so (\textit{optimal control problem}). In control theory this is largely solved, and we will now consider how to combine the resulting solutions with SCN theory to generate the correct $\textbf{u}$ as an output of an SNN.

\subsection{Optimal state estimation with SCNs}\label{Sec_SCN_KF}
Full-state estimates $\B{\hat{x}}$ given a noisy and incomplete measurements in linear systems can be provided by a Kalman filter~\cite{brunton_data-driven_2019}, which optimally balances an internal (predicted) state estimate to a noisy and/or partially observable external measurement (Fig.~\ref{fig:1}B.i). The Kalman filter in continuous time is given by a dynamical system of the form
\begin{equation}\label{KF_ds}
    \dot{\hat{\B x}} = \B A \hat{\B x} + \B B \B u + \B K_{f} (\hat{\B y} - \B y) \text{.}
\end{equation}
The Kalman filter gain matrix, $\B{K}_{f} \in \RR^{K \times Q}$, is applied to an error between the observations of the dynamical system and the Kalman filter's own internal estimate, $\B{\hat{y} = C\hat{x}}$. For fully identified systems $\B{K}_{f}$ can be found by solving an algebraic Riccati equation \cite{brunton_data-driven_2019}. Assuming that $\B{K}_{f}$ is known, we can directly implement the optimal filtering under the SCN framework (by following Section~\ref{sec:dynimpl}) resulting in the following voltage update rule:
\begin{align}\label{derivation_with_kf}
    \B{\dot{v}} &= -\lambda\B{v} + \underbrace{\B{\Omega}_s\B{r} + \B{\Omega}_f\B{s}+ \B{F}_i\B{u}}_{\text{System estimate}}  + \underbrace{\B{\Omega}_k\B{r} + \B{F}_k\B{y}}_{\text{Kalman update}} + \eta_{V} \text{,}
\end{align}
where, on top of the previously introduced slow and fast connectivity, we now also have input (control) connections $\B{F}_i=\B{D}^\top\B{B}$ mapping the control signal $\B u$ to the neurons, recurrent  and feed-forward ``Kalman filter" connections ($\B{\Omega}_{k} = -\B{D}^\top\B{K}_{f}\B{CD}$ and  $\B{F}_k = \B{D}^\top\B{K}_{f}$). The recurrent connections essentially read-out the internal state of the system, apply both the observability matrix and the Kalman filter gain matrix to predict its evolution, and maps this back into the network. The feed-forward connections take the partially observable state, $\B{y}$, and applies the Kalman filter gain matrix. A more detailed derivation can be found in the appendix (section \ref{Sec:Appendix}). Now, the entire SCN, including all of its connections, represents a Kalman filter (Fig.~\ref{fig:1}B.ii). The network internally simulates a linear dynamical system of the form $\B{\dot{\hat{x}} = A\hat{x} + Bu}$, but also corrects its own estimate according to the external input $\B{y}$. 

\subsection{Optimal control with SCNs}\label{Sec_SCN_KF_and_LQR}
 Now that we have optimal state-estimation we can extend the framework to optimal control by deriving the spiking version of a Linear-Quadratic Regulator (LQR) controller. Given a known system of the form in Eq.~\eqref{eq:dynsystem_x}, this controller produces the optimal $\B u$ signal to minimize the squared error between the state $\B x$ and a target $\B z$. This gives rise to a linear control law of the form
\begin{equation}\label{LQR_controllaw}
    \B{u} = -\B{K}_{c}(\B{x}-\B{z}) \text{,}
\end{equation}
where $\B{K_{c}} \in \RR^{P \times K}$ is the LQR gain matrix, which can be found by solving an algebraic Riccati equation, given assumptions on the cost of state deviations and actuation \cite{brunton_data-driven_2019}. 

In order to control a partially observable and/or noisy dynamical system, we can combine an LQR controller with a Kalman filter (see Fig.~\ref{fig:1}C.i). The Kalman filter should then be aware of the specific form used for the control, and we can swap out $\B{u}$ for the LQR control law (Eq.~\eqref{LQR_controllaw} in Eq. \eqref{KF_ds}), resulting in a dynamical system of the form 
\begin{equation}\label{Eq:Kalman_LQR}
   \B{\dot{\hat{x}} = A\hat{x} + B}\B{K}_{c}(\B{\hat{x}} - \B{z}) + \B{K}_{f}\B{(\hat{y} - y).} 
\end{equation}
We will refer throughout to the combination of a Kalman filter and LQR control as the \textit{ideal controller}. We can now use the same method as in the previous section to implement this in a spiking network.

To be able to read out the full control signal, we additionally encode the reference $\B{z}$ into the network with a new set of fast connections (similar to Eq. \eqref{derivation_encoder}), which allows us to read out $\B{u}$ from the internal estimates $\B{\hat{x}}$ and $\B{\hat{z}}$ of the network. For this we define two sets of decoding weights: $\B{D}_\B{x}$ for $\B{\hat{x}}$, and $\B{D}_\B{z}$ for $\B{\hat{z}}$. 

Extending the SCN defined in Eq. \eqref{derivation_with_kf} with LQR control and an internal encoding of $\B{z}$, we get the final voltage update rule
\begin{align}\label{derivation_with_kfandlqr}
\begin{aligned}
        \B{\dot{v}} = -\lambda\B{v} &+ \underbrace{\B{\Omega}_s\B{r} + \B{\Omega}^\B{x}_f\B{s}}_{\text{System estimate}} + \underbrace{\B{\Omega}_c\B{r} + \B{\Omega}_{z}\B{r}}_{\text{Control estimate}}+ \underbrace{\B{\Omega}_k\B{r} + \B{F}_k\B{y}}_{\text{Kalman update}} \\
        &+ \underbrace{\B{D}_\B{z}^\top(\B{\dot{z}} + \lambda\B{z}) + \B{\Omega}^\B{z}_f\B{s}}_{\B{z} \text{ representation} } + \eta_{V} \text{,}
    \end{aligned}
\end{align}
where we have indicated the parts of the connectivity that are tracking the external system, the effect of the control on the system, the Kalman filter updates, and the representation of the target signal.  For the internal estimate $\B{\hat{x}}$, we now have two sets of recurrent ``control" and ``target" connections, which represent the LQR controller within the Kalman filter SCN (see Fig.~\ref{fig:1}C.ii). The recurrent control connectivity, $\B{\Omega}_{c} = -\B{D}_\B{x}^\top\B{B}\B{K}_{c}\B{D}_\B{x}$, decodes $\B{\hat{x}}$ from the internal state of the SCN using $\B{D}_\B{x}$, and applies the LQR gain matrix $\B{K}_{c}$. The result is transformed using $\B{B}$, and encoded back into the network. The recurrent target connectivity, $\B{\Omega}_z = \B{D}_\B{x}^\top\B{B}\B{K}_{c}\B{D}_\B{z}$ does something similar, but first reads out $\B{\hat{z}}$ using $\B{D}_\B{z}$. The reference signal, $\B{z}$, is encoded into the network using the same method as defined in Eq. \eqref{derivation_with_kf}. A more detailed derivation for all these connections can be found in the appendix (section \ref{Sec:Appendix}). Note that while there now appear to be a large set of separate recurrent connections, all the slow connectivities ($\B{\Omega}_{s}$, $\B{\Omega}_{c}$, $\B{\Omega}_{z}$, $\B{\Omega}_{k}$) can be grouped together in a single connection matrix, as well as both the fast connectivities ($\B{\Omega}_{f}$, $\B{\Omega}^\B{z}_{f}$).

We now have an SCN which implements the entirety of $\B{\dot{\hat{x}} = A\hat{x} + B}\B{K}_{c}(\B{\hat{x}} - \B{\hat{z}}) + \B{K}_{f}\B{(\hat{y} - y)}$, but also internally keeps an estimate of the desired state or reference $\B{z}$. We can obtain the output of the internal LQR controller in the SCN by defining the following new set of decoding weights: $\B{D_u} = -\B{K}_{c} (\B{D}_\B{x} - \B{D}_\B{z})$. The control signal can then directly be read-out from the neural activities as $\B{u} = \B{D_u}\B{r}$. This $\B{u}$ can be applied to an external dynamical system to control it, all whilst the SCN keeps an internal estimate of the external dynamical system. Hence, we have finally arrived at a fully derived recurrent SNN which allows us to track, replicate and control an external dynamical system.

\section{Results}
We evaluated the performance of the proposed mathematical framework (Sec.~\ref{sec:methods:model}) as well as analyzed the properties of the spiking patterns for internal robustness against perturbations (e.g., noise, neural silencing and external force)\footnote{The code used to generate the results is available in the following Github repository: \url{https://github.com/FSSlijkhuis/SCN_estimation_and_control}}. We applied the networks we derived above to two standard dynamical systems: the linear spring-mass-damper system and the nonlinear cartpole system.

\subsection{Spring-Mass-Damper (SMD) system}\label{sec:SMD-results}
\begin{figure*}[ht]
\vspace{-1em}
    \centering
    \includegraphics[width=0.8\textwidth]{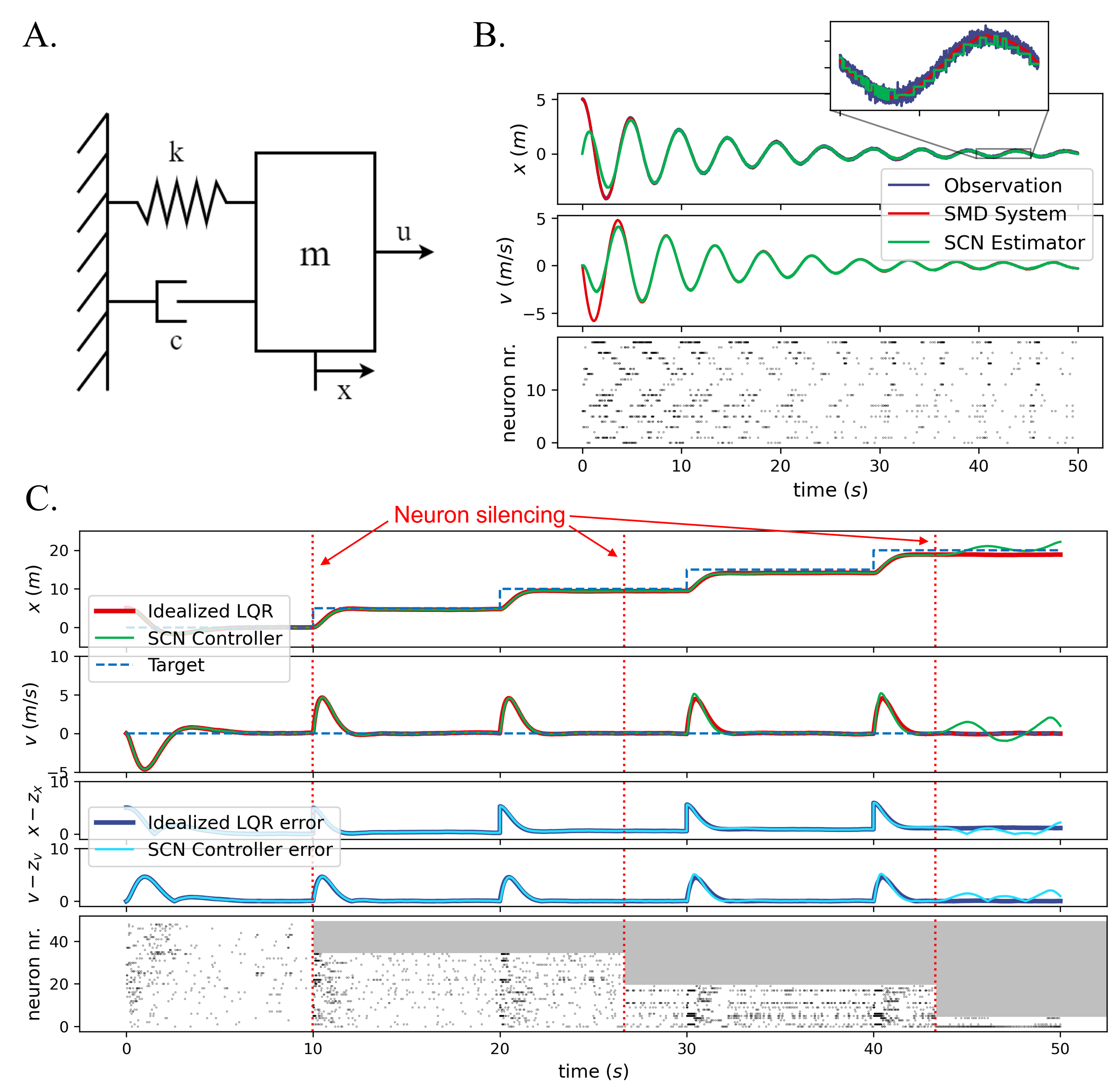}
    \caption{Estimation and control of a linear system using SCNs. \textbf{(A)} Schematic of a Spring-Mass-Damper (SMD) system. \textbf{(B)} SCN estimation \textit{(green)} of SMD system \textit{(red)} simulated over a time period of 50 seconds. The top and middle plot show the estimated and real positions of the spring and the estimated and real velocity of the spring in the SMD system, respectively. The observation (\textit{blue}) is shown in the top plot only, since the SCN estimator only has access to the noisy $x$. The bottom plot shows the spike trains across all 20 neurons used in the SCN estimator. \textbf{(C)} Control of SMD system using the SCN controller \textit{(green)} vs  an idealized controller \textit{(red)}, with reference signal \textit{(blue, dashed)}, while evaluating neural silencing robustness. The reference position of the spring is increased in a stair-wise manner, which explains the sudden changes in the spring velocity. Increased spiking activity in the neurons of the SCN is observed during these shifts. The robustness to neural silencing of the SCN controller is also shown. At every red, dashed vertical line, 15 neurons are ``killed" (indicated by the gray bars in the spike plot). The SCN controller and estimator starts with 50 neurons, and is able to correctly control and estimate the SMD system up until the last portion of neurons is killed, after which the system breaks down due to control errors. Before the system fails, the SCN shows some signs of neurons 'taking over' the spiking of removed neurons. In the third and fourth plot, the error between the target signal and both controllers is shown for both the position and the velocity of the spring.}
    \label{fig:SMD-results}
\vspace{-1em}
\end{figure*}

We compared our proposed SCN estimation and control with their non spiking counterparts for the spring-mass-damper system (Fig.~\ref{fig:SMD-results}A). The baselines comparison are a Kalman filter for estimation and LQG for control, which we refer to as the "Idealized" estimator and controller. Furthermore, we evaluated the system robustness against input-noise, neural silencing and external perturbations. The SMD system has two dynamical variables: position $x$ ($= x_{1}$) and velocity $v$ ($= x_{2} = \dot{x}_{1}$). Our instance of the SMD system has an internal disturbance $\B{\eta}_{d}$, and it outputs a partially observable state which measures only the position $x$ with sensor noise $\B{\eta}_{n}$.

\subsubsection{Estimation}
We first evaluated whether an SCN Kalman filter can accurately track the SMD system based on incomplete and noisy measurements. For the estimation, we set $u=0$, so there is no external force acting on our dynamical system. In Fig.~\ref{fig:SMD-results}B, we compared the estimation outputted by the spike coding network (green) to the numerical dynamical system simulation (red). Even though the estimation starts in a different initial state and there is both substantial measurement noise and internal disturbance in the real dynamical system, the network estimate quickly converges. Overall, the SCN estimates both dynamical variables of the real dynamical system with very high accuracy.



\subsubsection{Control}
To control a dynamical system, the SCN computes the optimal $u$, such that the state of the simulated dynamical system converges towards a reference signal $\B{z}$. In Fig.~\ref{fig:SMD-results}C, we show an SMD controlled through an idealized controller (red) and a dynamical system controlled using an SCN controller (green). Once again, the estimation from the SCN is very similar to the real dynamical system. On top of that, we show the reference signal (blue, dashed), which is a stair-wise increase of the SMD position, $x$. We can see that both the estimation from the SCN and the real dynamical system follow the reference signal, which is an indication of the correct behavior of the internal controller of the SCN. When we investigate the control errors in detail (blue curves) we see that the two controllers match closely. 


\begin{figure*}[ht]
\vspace{-1em}
    \centering
    \includegraphics[width=0.8\textwidth]{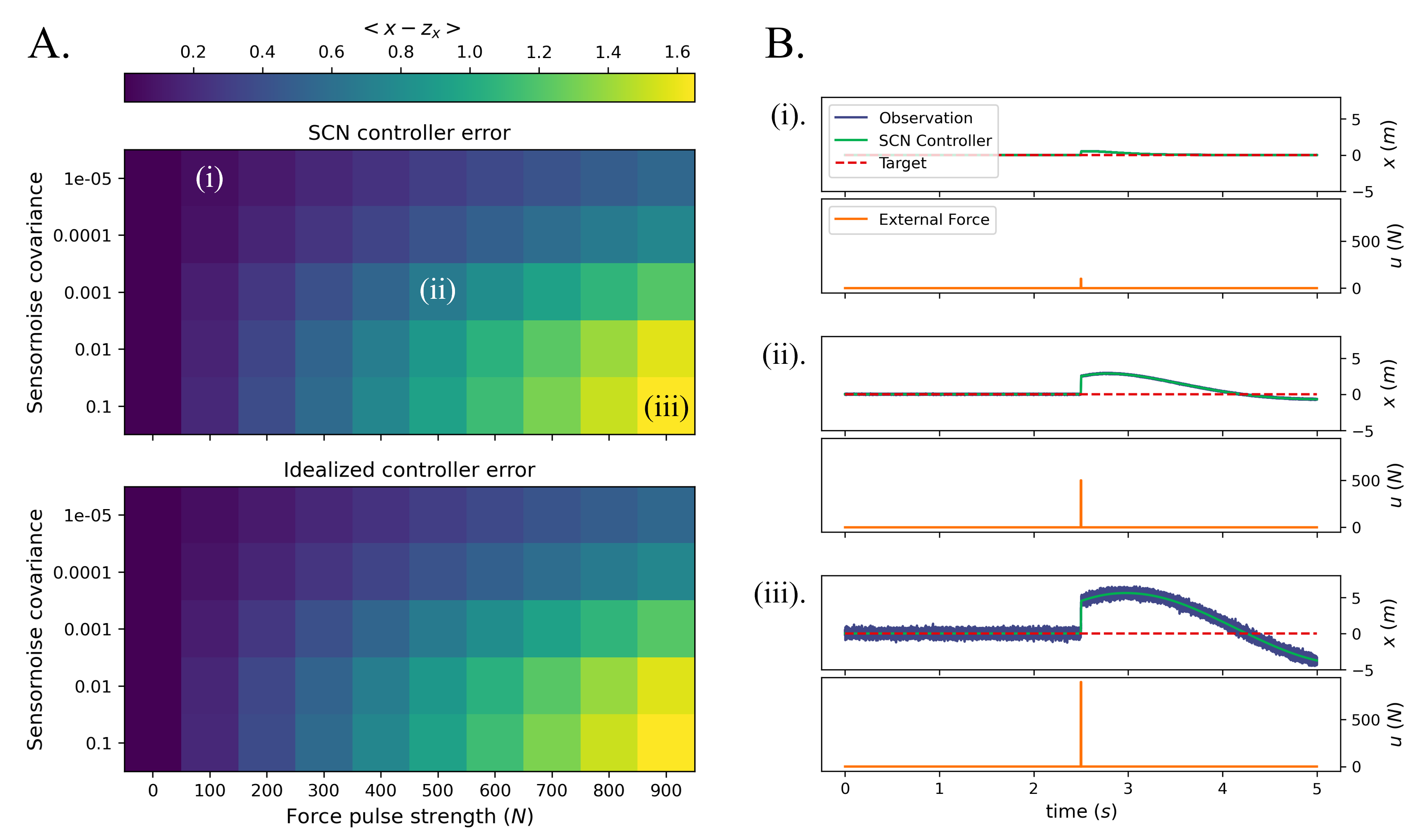}
    \caption{Spring-Mass-Damper system control range of SCN controller. \textbf{(A)} Average position error between SMD system controlled through SCN controller and target signal (top), where a short force pulse is applied halfway during the simulation, plotted against multiple values for the sensor noise covariance. The same analysis is shown for the idealized controller (bottom). Both the sensor noise covariance and the strength of the force pulse are varied, and for each combination of values, a simulation of 5 seconds is run. The result is shown as the error in the position of the SMD system averaged over the entire simulation, where dark blue indicates a small error, and yellow a large error. \textbf{(B)} The effect of force pulses and sensor noise on the SMD system when controlled by an SCN controller. The influence on the SMD system is shown for three parameter values, which are highlighted in A. The top plot, (i), shows the effects of a noise covariance of 1e-5 and a force pulse strength of 100N. Plotted are the (noisy) observations going into the SCN controller \textit{(blue)}, the SMD system controlled by the SCN controller \textit{(green)}, and the reference signal \textit{(red, dashed)}. Plotted separately is the pulse into the SMD system (\textit{orange}). The middle plot, (ii), shows the same but with a sensor noise covariance of 0.001 and a pulse strength of 500N. The bottom plot, (iii), uses a sensor noise covariance of 0.1 and a pulse strength of 900N.}
    \label{fig:SMDcontrolrange-results}
\vspace{-1em}
\end{figure*}

\subsubsection{Controller robustness}

In Fig.~\ref{fig:SMD-results}C, we further show the robustness of our framework, as the network is able to keep the system controlled when facing severe neural silencing (as in classic SCNs \cite{barrett_david_optimal_2016,calaim_geometry_2022}). Starting with 50 neurons, we progressively disable 15 neurons at certain timesteps (red, vertical dashed lines),  by preventing these neurons to spike from this point on.  The timestep at which the first 15 neurons are disabled (at 10 seconds) is exactly at the point of a change in the reference signal.  In this case there is no visible effect on the performance. The neural silencing at the second timestep (at 26.6 seconds) again demonstrates little effect on the performance of the SCN controller, although a very subtle increase in the velocity error is observed. After the third timestep at which 15 neurons are disabled (at 43.3 seconds), only 5 neurons are available to the SCN controller. Only at this point does the SCN controller start to struggle to accurately control the SMD system. 


In the raster plots in Fig.~\ref{fig:SMD-results}, the spike patterns are moderately sparse and irregular. For clarity of the spike trains we used smaller networks here, but the spiking patterns can be made arbitrarily sparse by adding more neurons (as then the spiking is coordinated across more neurons). In Fig. \ref{fig:SMD-results}C, we can observe that when neurons are silenced, other neurons compensate for its spiking, to make sure that the estimation stays as accurate as possible (and the spikes become progressively less sparse). This is a clear indication of the coordinated spiking of the neurons in the network.

Fig.~\ref{fig:SMDcontrolrange-results}A shows the effects of different parameter-values for the sensor noise covariance and the effects of a sudden pulse of force into the SMD system on the controller error of the SCN controller (top) and the idealized LQR controller (bottom). Here, the controller error corresponds to the average error in the position of the mass between the SMD system and the reference signal across an entire simulation, when controlled by the respective controllers. This error increases for larger values of the sensor noise covariance, as well as with larger pulses of force into the SMD system. There is no noticeable difference between the control errors of the SCN controller and the idealized controller. 

In Fig.~\ref{fig:SMDcontrolrange-results}B, the position of the mass in the SMD system controlled by the SCN controller is shown, using three sets of parameters indicated in Fig.~\ref{fig:SMDcontrolrange-results}A. Here, we clearly see the effect of the pulse on the SMD system, where a larger pulse corresponds to a larger displacement of the mass. The SCN controller quickly tries to correct for this displacement, bringing the mass back to its reference position. The observation, $\textbf{y}$, is also shown. Larger values for the sensor noise covariance are clearly visible in the observation. Not shown here is the SMD system controlled by the idealized controller, because there is no visible difference between the two SMD systems. 
\begin{figure*}[ht]
    \centering
    \includegraphics[width=0.8\textwidth]{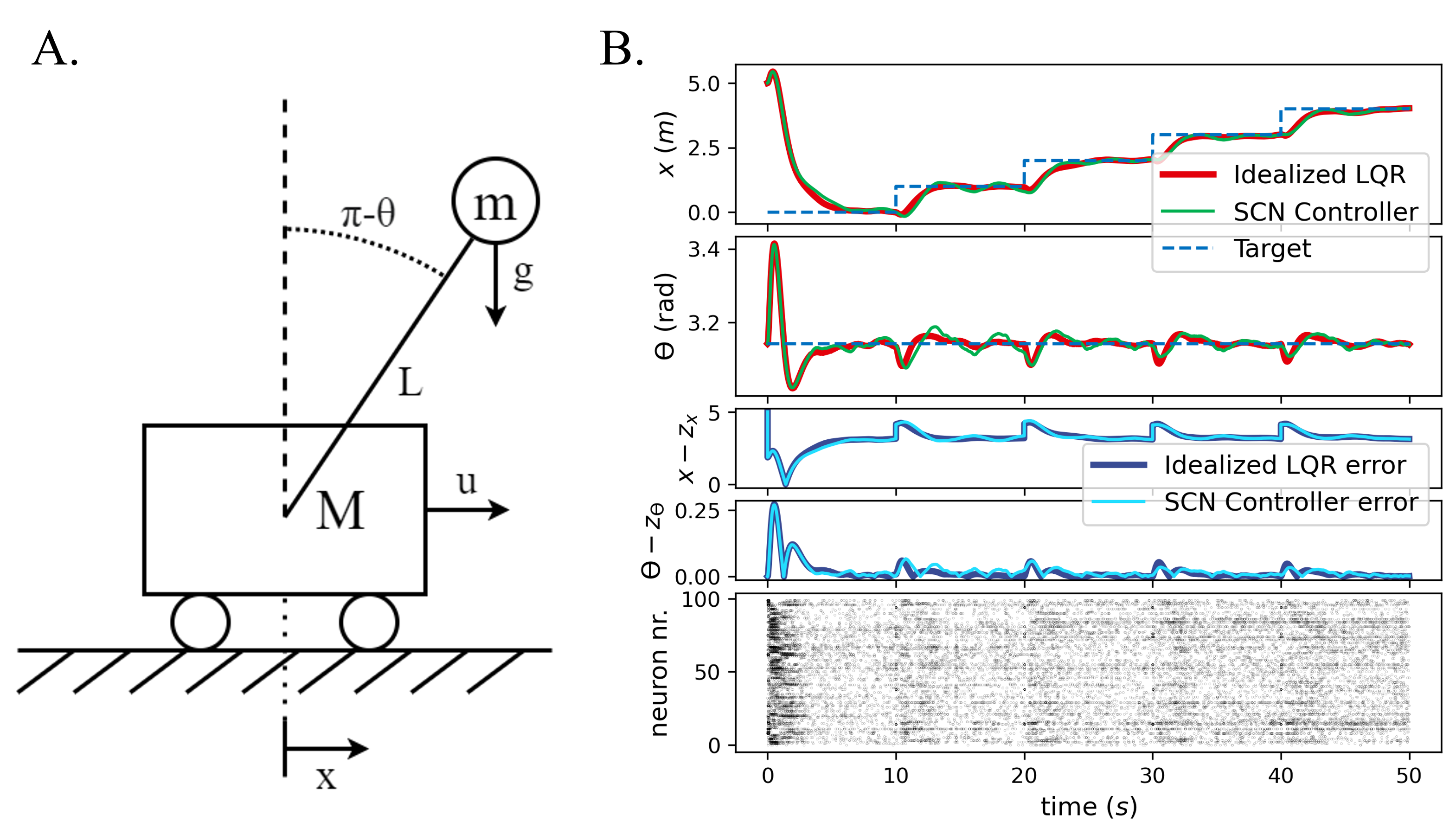}
    \caption{SCN control of nonlinear cartpole system. \textbf{(A)} Schematic of the Cartpole system. \textbf{(B)} Control of Cartpole system \textit{(green)} using SCN controller and estimator linearized around up-position of pole, compared to a second Cartpole system \textit{(red)} using idealized linear controller and estimator, also linearized around the up-position of the pole, with reference signal \textit{(blue, dashed)}, simulated across a time period of 50 seconds. The figure only shows the position of the cart ($x_{1}$) and the angle of the pole ($x_{3}$). The reference position of the cartpole is increased in a stair-wise manner, but the pole stays upright. In the third and fourth plot, an error between the target and both controllers is shown, for both the position of the cart and the angle of the pole. The lower plot shows the spiking activity within the SCN, consisting of 100 neurons. During a change in the reference position, an increase in spiking activity is observed.}
    \label{fig:cartpole-results}
\vspace{-1em}
\end{figure*}

\subsection{Cartpole system}

We evaluated our proposed SCN controller (i.e., linear-quadratic-Gaussian spiking control) on the nonlinear cartpole system (see Fig.~\ref{fig:cartpole-results}A). The cartpole has four dynamical variables: the position of the cart, $x$ ($= x_{1}$), the velocity of the cart $v$ ($= x_{2} =  \dot{x}_{1}$), the position of the pole, $\theta$ ($= x_{3}$), and the angular velocity of the pole, $\omega$ ($= x_{4} =  \dot{x}_{3}$). Just like the SMD, our instance of the cartpole has an internal disturbance $\B{\eta}_{d}$, and it outputs a partially observable state consisting of only the position of the cart, $x$, with sensor noise $\B{\eta}_{n}$. Note that the cartpole system is a nonlinear dynamical system, but both our ideal and spiking controllers assume linear dynamics. To produce a meaningful control signal, we use linearized dynamics for the internal estimate. In our case, we linearized around the up-position of the pole ($\theta = \pi$).

Fig.~\ref{fig:cartpole-results}B compares a cartpole system controlled through an idealized controller (red) to a cartpole controlled through an SCN controller (green). On top of that, we show the reference signal (blue, dashed), which is a stair-wise increase of the cart position, $x$. The task of the controllers is then to move the cart whilst keeping the pole upright. In the figure, we show the position ($x$) of the cart and the angle of the pole ($\theta$). We see that both controllers follow the reference signal almost perfectly, and most importantly, keep the pole upright. 

The spike plots show that especially with a large amount of neurons, the spiking is irregular and sparse. Increased spiking activity is observed when the reference signal demands a change of the state of the dynamical system.
%


\subsection{Parameter settings}\label{sec:methods:par}
For each network, the decoding weights ($\B{D}$) are sampled from a normal distribution, with each column normalized to have norm 0.1, except for norm 0.01 in Fig.~\ref{fig:cartpole-results}B and norm 1 in Fig.~\ref{fig:SMDcontrolsparsity}. 20 and 50 neurons were used in  Fig.~\ref{fig:SMD-results}B and C respectively, 50 neurons were used in Fig.~\ref{fig:SMDcontrolrange-results} and Fig.~\ref{fig:SMDcontrolsparsity}, and 100 neurons in Fig.~\ref{fig:cartpole-results}B. The SMD parameters were $m=3$, $k=5$ and $c=0.5$ for Fig.~\ref{fig:SMD-results}B and Fig.~\ref{fig:SMDcontrolrange-results}, while $m=20$, $k=6$ and $c=2$ were used for Fig.~\ref{fig:SMD-results}C and Fig.~\ref{fig:SMDcontrolsparsity}. Fig.~\ref{fig:SMD-results}B and Fig.~\ref{fig:SMDcontrolsparsity} used $\B{\Sigma}_{d} = \B{\Sigma}_{n} = 0.001$. Fig.~\ref{fig:SMD-results}C used $\B{\Sigma}_{d} = \B{\Sigma}_{n} = 0.1$. Fig.~\ref{fig:SMDcontrolrange-results} used $\B{\Sigma}_{d} = 0.001$. The cartpole system parameters were $m=1$, $M=5$, $L=2$, $g=-10$, and $d=1$, with $\B{\Sigma}_{d} = \B{\Sigma}_{n} = 1\text{e-}7$. All networks used $\eta_{V} = 1\text{e-}5$ and $\lambda = 0.1$, except for Fig.~\ref{fig:SMDcontrolsparsity}, which used $\eta_{V} = 1\text{e-}6$ and varying $\lambda$. Forward Euler was used throughout, with a timestep of 1e-3s, except for 1e-4s in Fig.~\ref{fig:SMDcontrolrange-results}, Fig.~\ref{fig:cartpole-results}B and Fig.~\ref{fig:SMDcontrolsparsity}. For control, $R = 1\text{e-}2$. For control of the SMD system, the matrix $\B Q \in \mathbb{R}^{K\times K}$ prioritizes $x_{1}$ with weight 10, and $x_{2}$ with weight 1. For cartpole control, $x_{3}$ is weighted with 10 and the rest with 1.

\section{Advantages and limitations of SCN control\label{sec:adlims}}
The use of SCN control provides several advantages and limitations. This section provides an overview.

\subsubsection{Hardware implementations}
Because SCN control is implemented through standard LIF neurons it has strong potential for neuromorphic deployment, such as on Loihi \cite{davies_loihi_2018} or SpiNNaker \cite{rhodes_real-time_2019} chips --- allowing for potentially lower energy control implementations than classical von Neumann architectures. However, a general framework of how to transfer the theory to hardware is still lacking, and faces two main challenges. First, the theory assumes zero transmission delays between neurons, which effectively constraints one neuron to spike per time-frame. If this constraint is weakened, the networks can go into an epileptic state \cite{calaim_geometry_2022,buxo_poisson_2020}. On hardware implementations this assumption might not always be reasonable. As a solution, several extensions have been made to SCN theory that make it work with such delays \cite{chalk_neural_2016,buxo_poisson_2020,calaim_geometry_2022,mikulasch_local_2021}. Second, for control we need two types of connections: fast and slow. The fast connections can be implemented by a simple change in the voltage following a spike, but the slow connections require longer lasting effects (through the filtered spike trains $\B r$). Not every hardware implementation might allow for such slow synapses, or allow the simultaneous use of both synapse types. It also requires every neuron to have access to the filtered spike trains of every other neuron, which might increase the connectivity requirements --- which is again highly dependent on the exact hardware.

\subsubsection{Explainability}
\begin{figure}[t]
\hspace{-1.3em}
    \includegraphics[width=1.03\linewidth]{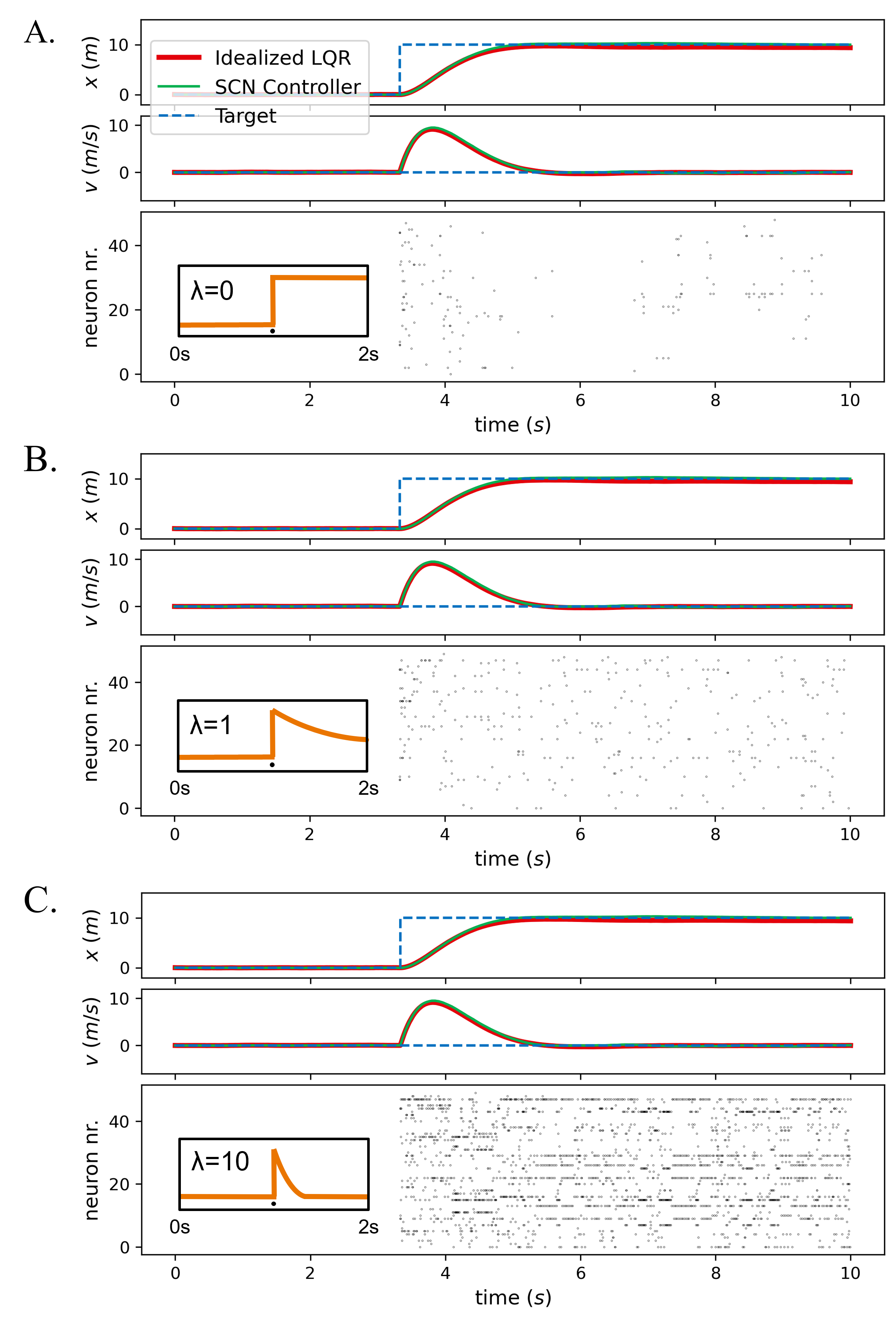}
    \caption{Demonstrating adjustment to the network sparsity through adjusting the voltage leakage ($\lambda$), shown on the Spring-Mass-Damper system, simulated over a time period of 10 seconds. For all values of $\lambda$, the normalization factor of the decoding weights has been increased to 1 in order to reduce overall spiking at the cost of controller precision. \textbf{(A)} The effects of $\lambda=0$ on spiking behavior of the SCN controller. A total of 163 spikes are fired during this simulation, which averages to 16 spikes per second. Inset: a filtered spike train $r_n$ for a single neuron $n$ after a single spike (dot). \textbf{(B)} The effects of $\lambda=1$ on spiking behavior of the SCN controller. A total of 358 spikes are fired during this simulation, which averages to 35 spikes per second. \textbf{(C)} The effects of $\lambda=10$ on spiking behavior of the SCN controller. A total of 2381 spikes are fired during this simulation, which averages to 238 spikes per second.}
    \label{fig:SMDcontrolsparsity}
\vspace{-1em}
\end{figure}

A great benefit of SCN control is its explainability. There is a clear link between the different network parameters and the computation (as outlined in our methods section), as well as a rich literature studying their more in-depth properties. In particular, there is an elegant and in-depth geometric interpretation of the core functionality \cite{calaim_geometry_2022}. Thus, with new applications for the theory and resulting problems, there is a clear theoretical framework within which one can think through possible solutions. One can also easily adjust the model parameters for different activity and coding regimes, as demonstrated in Fig.~\ref{fig:SMDcontrolsparsity}. Another spiking neural network method that has a similar level of theoretical explainability is the neural engineering framework \cite{eliasmith_large-scale_2012}, which also allows one to implement dynamical systems, including for control\cite{dewolf_spiking_2016}. However, as outlined next, SCNs come with a number of additional advantages --- in particular for robustness and biological fidelity.

\subsubsection{Biological fidelity}
A scientifically motivated advantage of SCN control over other methods of control is its biological plausibility. Despite being directly derivable through the underlying theory, the resulting networks portray several features consistent with biology, such as sparse and irregular firing \cite{tolhurst_statistical_1983,shadlen_variable_1998}, robustness to a range of perturbations\cite{barrett_david_optimal_2016, calaim_geometry_2022}, and an underlying strict balance of excitation and inhibition\cite{boerlin_predictive_2013, deneve_efficient_2016}. The theory is also extendable to more complex synapses and neural models either through learning\cite{mikulasch_local_2021} or design\cite{nardin_nonlinear_2021}, enabling more complex computations. 

\subsubsection{Distributed and robust control}
A core advantage of SCN control is that we inherit the property of previous SCN implementations of distributed and robust computations\cite{boerlin_predictive_2013, barrett_david_optimal_2016}. Given a known linear control algorithm, the required computations can be effectively distributed across the neurons. Consequently, the resulting networks are highly robust to neural silencing (Fig.~\ref{fig:SMD-results}). The robustness to neural silencing might be highly useful for situations in which neuromorphic chips can get damaged during operation, such as in high radiation environments.

\subsubsection{Temporal sparsity and energy use}
The distributed and coordinated nature of SCNs ensures that spikes are only fired if they contribute to the required dynamics. As a result, the output spike patterns can be highly temporally sparse and efficient. Compared to other spiking neural network methods, SCNs therefore might potentially offer lower energy use. However, for a fair comparison to other methods we would need direct hardware implementations and comparison to other methods, and measure the energy directly. While this is out of scope for this paper, we here highlight a few important caveats. First, while the total spike count across the population is highly optimized, this does require highly dense recurrent connectivity, and thus significant spike-routing and potential energy use. While the density of connectivities can be drastically reduced\cite{nardin_nonlinear_2021}, it would still need to be taken into account. Second, the use of both fast and slow connectivity types might complicate neural interactions, further increasing energy use. Finally, the sparsity might not be as optimized in the presence of synaptic delays. Even with these caveats, the theoretical understanding of the method should allow further optimization once comparisons can be made. As a demonstration, we show one can drastically reduce the total number of spikes by reducing the voltage leakage constant --- without losing performance or needing retraining (Fig.~\ref{fig:SMDcontrolsparsity}) (at the cost of less biological realism and potentially higher memory requirements).

\section{Conclusion}
For fully identified systems there exist highly efficient closed-form control solutions. Ideally we would like to be able to implement these directly in spiking neural networks, and in a way that mimics the efficient and robust nature of the brain. In this work, we provided an extension of SCN theory to allow optimal estimation and control --- providing spiking equivalents for Kalman filters and LQR control. We showed that these networks maintain sparse and irregular spiking patterns when controlling a dynamical system, and are robust to both severe neural silencing and system perturbations. Our networks are analytically derived and do not need learning nor optimization. Hence, our presented approach opens up the prospect for deploying fast, efficient and low-power SNN on-chip controllers with the advantage of having hardware intrinsic redundancy (i.e., the controller still works when some of the neurons stop working) and ensuring that neurons only fire when there is a prediction error.

As outlined in section \ref{sec:adlims}, there are several open challenges for future work, in particular for hardware implementations. Of particular note is that SCN theory assumes instantaneous synaptic delays, which is not realistic in either the brain or all neuromorphic hardware. There are several avenues to implement delays in SCNs~\cite{chalk_neural_2016,buxo_poisson_2020,calaim_geometry_2022}, which can be well combined with the work presented in this paper. While here we focused on analytical implementations, learning rules do exist to implement SCN connectivities \cite{alemi_learning_2018, brendel_learning_2020,buchel_supervised_2021}. It is an open question whether these can also be applied for optimal control. Of particular difficulty then is online estimation of the Kalman and LQR gain matrices ($\B K_c$ and $\B K_f$). This was recently shown to be possible in non-spiking networks~\cite{friedrich_neural_2021}, which gives a possible avenue for combining online gains optimization within a closed-form spiking implementation.

\section{Acknowledgements}
We thank Marcel van Gerven, Bodo R\"uckauer, Justus H\"ubotter, Christian Machens, William Podlaski, and Michele Nardin for helpful discussions on control and (spiking) networks. 
\section{Appendix: detailed derivation of network connectivity}\label{Sec:Appendix}
Here we will show how one can use SCN theory to take the dynamical system corresponding to the continuous Kalman filter and LQR control, and directly implement them in an SNN (following \cite{boerlin_predictive_2013}). For both cases we start with the auto-encoding network
\begin{equation}
    \B{\dot{v}} = -\lambda\B{v} + \B{D}^\top(\B{\dot{x}} + \lambda\B{x}) + \boldsymbol \Omega_f \B s \text{.}
\end{equation}
This network takes as an external input $\B x$ and its derivative $\dot{\B x}$ and can accurately track this signal with the read-out $\B D \B r$. To implement a given dynamical system in this network we need to do two things. (1) replace the derivative of the input by the desired dynamics ($\dot{\B x} = f(\B x, \B u)$). (2) Instead of feeding the state of $\B x$ as an external input, estimate the current state from the network read-out $\hat{\B x} = \B D \B r$, and feed that back into the input (by simply replacing $\B x$ by $\hat{\B x}$). This results in network dynamics given by:
\begin{equation}
    \B{\dot{v}} = -\lambda\B{v} + \B{D}^\top(f(\hat{\B x}, \B u) + \lambda \hat{\B x}) + \boldsymbol \Omega_f \B s \text{.}
\end{equation}

If $f(\B x, \B u)$ is purely linear this can be done through a set of slow connections, as in the main text. We do not consider nonlinear cases in this paper, but note that an analytical solution does exist\cite{nardin_nonlinear_2021}.

\subsection{Estimation}
For the continuous definition of the Kalman filter (Eq.~\ref{KF_ds}), the above results in voltage dynamics given by

\begin{align}
\begin{aligned}
\B{\dot{v}} = -\lambda\B{v} &+ \B{D}^\top(\B{A\hat{x}} + \lambda\B{\hat{x}}) - \B{D}^\top\B{Ds} + \B{D}^\top\B{Bu}\\
 &  + \B{D}^\top \B{K}_{f} (\B{\hat{y}} - \B{y}) + \eta_{V}.
\end{aligned}
\end{align}
We can then replace $\hat{\B x} = \B D \B r$, and group the different terms together to get the final voltage dynamics to implement a Kalman filter in a recurrent SNN given by Eq.~\ref{derivation_with_kf}.

\subsection{Control + estimation}
For the continuous definition of a Kalman filter including LQR control we have equation~\ref{Eq:Kalman_LQR}. We can do the same as above and directly implement this in a recurrent SNN, but we also additionally need to represent the target state $\B z$ in the network (as explained in the main text). This together results in the voltage update rule

\begin{align}
    &\begin{aligned}
        \B{\dot{v}} = -\lambda\B{v} &+ \B{\Omega}_s\B{r} - \B{D}_\B{x}^\top\B{D}_\B{x}\B{s} + \B{D}_\B{x}^\top\B{B}\B{K}_{c}(\B{\hat{x}} - \B{\hat{z}})  \\ 
        &+ \B{D}_\B{x}^\top \B{K}_{f} (\B{\hat{y}}- \B{y})+ \B{D}_\B{z}^\top(\B{\dot{z}}+ \lambda\B{z}) \\
        & - \B{D}_\B{z}^\top\B{D}_\B{z}\B{s} + \eta_{V}.
    \end{aligned}
\end{align}

We can then replace $\hat{\B x} = \B D \B r$, and group the different terms together the get the final voltage dynamics to implement a Kalman filter plus LQR controller in a recurrent SNN given by Eq.~\ref{derivation_with_kfandlqr}.
%
%
%
%


\bibliographystyle{IEEEtran}
\bibliography{references}

\end{document}